%% file: camera.tex
\documentclass[10pt,twocolumn,letterpaper]{article}

\usepackage{cvpr}              
\usepackage{multirow}
\usepackage{algorithm}
\usepackage{algpseudocode}
\usepackage[table]{xcolor}

\input{preamble}
\definecolor{cvprblue}{rgb}{0.21,0.49,0.74}
\usepackage[pagebackref,breaklinks,colorlinks,allcolors=cvprblue]{hyperref}

\def\paperID{31553} 
\def\confName{CVPR}
\def\confYear{2026}

\title{ShiftLUT: Spatial Shift Enhanced Look-Up Tables \\ 
for Efficient Image Restoration}

\author{
    Xiaolong Zeng$^{1}$\thanks{Equal contribution.} \qquad 
    Yitong Yu$^{2}$\footnotemark[1] \qquad 
    Shiyao Xiong$^{2}$\thanks{Project Lead.} \qquad 
    Jinhua Hao$^{2}$ \\
    Ming Sun$^{2}$ \qquad 
    Chao Zhou$^{2}$ \qquad 
    Bin Wang$^{1}$\thanks{Corresponding author. Email: wangbins@tsinghua.edu.cn} \\
    $^{1}$Tsinghua University, Beijing \qquad $^{2}$Kuaishou Technology, Beijing
}
\begin{document}
\maketitle
\input{sec/0_abstract}  \vspace{-10pt}
\input{sec/1_intro}
\input{sec/2_relatedwork}

\input{sec/3_method}

\input{sec/4_experiments}

\input{sec/5_conclusion}

{
    \small
    \bibliographystyle{ieeenat_fullname}
    \bibliography{main}
}
\input{sec/X_suppl}
\end{document}

%% file: sec/0_abstract.tex
\begin{abstract}
Look-Up Table based methods have emerged as a promising direction for efficient image restoration tasks. 
Recent LUT-based methods focus on improving their performance by expanding the receptive field. However, they inevitably introduce extra computational and storage overhead, which hinders their deployment in edge devices.
To address this issue, we propose ShiftLUT, a novel framework that attains the largest receptive field among all LUT-based methods while maintaining high efficiency. Our key insight lies in three complementary components. 
First, Learnable Spatial Shift module (LSS) is introduced to expand the receptive field by applying learnable, channel-wise spatial offsets on feature maps. 
Second, we propose an asymmetric dual-branch architecture that allocates more computation to the information-dense branch, substantially reducing inference latency without compromising restoration quality.
Finally, we incorporate a feature-level LUT compression strategy called Error-bounded Adaptive Sampling (EAS) to minimize the storage overhead.
Compared to the previous state-of-the-art method TinyLUT, ShiftLUT achieves a 3.8$\times$ larger receptive field and improves an average PSNR by over 0.21 dB across multiple standard benchmarks, while maintaining a small storage size and inference time.
The code is available at: \url{https://github.com/Sailor-t/ShiftLUT}.

\end{abstract}

%% file: sec/1_intro.tex
\section{Introduction}
\label{sec:intro}

With the widespread use of imaging systems on resource-constrained platforms such as smartphones and IoT devices, the demand for efficient and high-quality image restoration continues to grow. Image restoration, including tasks such as super-resolution, denoising, and deblocking, aims to reconstruct High-Quality (HQ) images from degraded Low-Quality (LQ) inputs. In recent years, Deep Neural Networks (DNNs) ~\cite{srcnn,vdsr,fsrcnn, hat, swinir, comparative, lei2024dvmsr} have achieved impressive performance in these tasks. However, their reliance on convolutional or transformer architectures incurs substantial computational overhead, thereby significantly limiting their deployment on resource-constrained devices.

\begin{figure}
    \centering
     \includegraphics[width=0.95\linewidth]
    {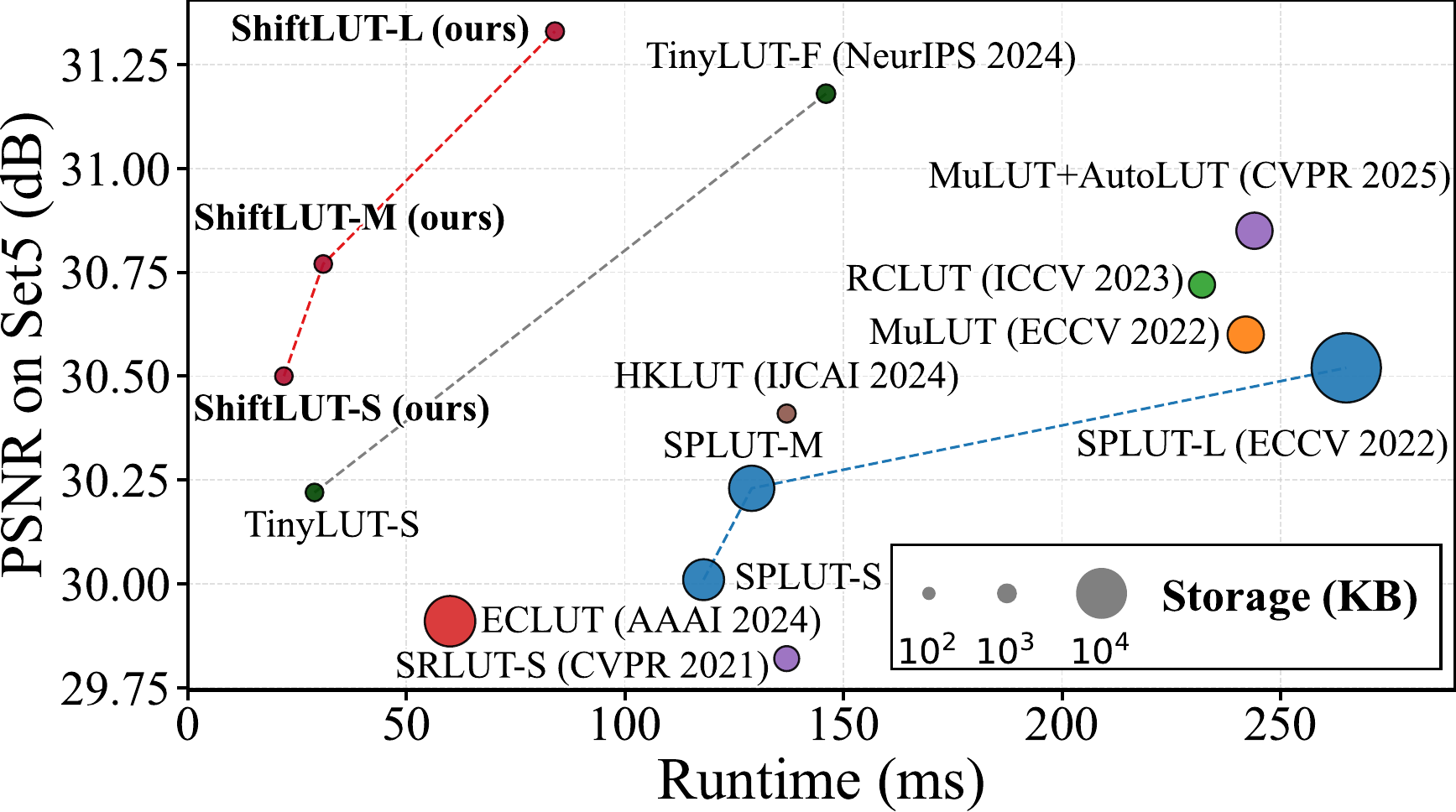}
    \caption{Model comparison in terms of storage size, PSNR and runtime on Set5 for x4 super-resolution. Our method produces a family of models that has the smallest storage size and occupies the top-left corner, indicating superior performances (PSNR on y-axis) with fast inference speed (Runtime on x-axis).}
    \label{fig:intro}
\end{figure}


To break this impasse, Look-Up Tables (LUT) based methods~\cite{srlut,mulut, splut, spflut, tinylut} have emerged as a new paradigm. By employing a ``space-for-time'' strategy, these methods replace expensive convolution operations with efficient memory lookups during inference, demonstrating potential for achieving low latency on edge devices. 
However, this efficiency comes at a cost: the receptive field is inherently constrained compared to those DNN-based methods, as expanding the receptive field leads to a significant increase in LUT computation and storage requirements. 
For instance, MuLUT~\cite{mulut} expands the receptive field by adopting multiple LUTs in a cascaded structure, yielding performance gains at the cost of increased storage and inference latency.

In this paper, we introduce \textbf{Learnable Spatial Shift module (LSS)}, a novel mechanism designed for LUTs to expand the receptive field with negligible computational and storage overhead. 
LSS introduces channel-wise spatial diversity into the LUT by learning a distinct spatial offset for each feature channel and applying the corresponding spatial shift to that channel.
This results in a significantly enlarged effective receptive field without modifying the LUT size, thereby effectively alleviating the storage and computation overhead that constrains other LUT-based approaches.


In addition to the challenge of receptive field expansion, we identify a critical inefficiency in the widely adopted dual-branch architecture, introduced by SPLUT~\cite{splut} to reduce LUT storage via decomposition of input signals into Most Significant Bits (MSB) and Least Significant Bits (LSB). Prior works~\cite{tinylut, taming} have predominantly employed symmetric designs, applying identical computational complexity to both branches. However, we argue that this symmetry is inefficient and conceptually flawed. 
Our empirical analysis shows that applying deep and complex LUTs to the LSB is inefficient, as the proportion of zero-valued activations in the LSB branch increases significantly
with depth.
Inspired by this observation, we propose an asymmetric architecture that simplify the LSB branch to a single LUT, effectively eliminating redundant computation. The saved resources are then reallocated to the information-rich MSB branch, yielding a more efficient and practical framework.


For LUT size compression, unlike previous sampling-based approaches~\cite{srlut, mulut, spflut} that employ a fixed stride across all LUTs, we introduce an \textbf{Error-bounded Adaptive Sampling (EAS)} algorithm that adaptively determines the optimal sampling stride for each LUT under a predefined error bound, effectively balancing reconstruction fidelity and compression efficiency. In addition, EAS introduces a lightweight caching mechanism that precomputes and stores intermediate interpolation results, eliminating redundant operations during inference, significantly accelerating runtime without additional memory cost.

By consolidating the aforementioned contributions, we propose ShiftLUT, a novel architecture for efficient image restoration. 
ShiftLUT substantially surpasses existing LUT-based methods in both restoration quality and efficiency.
Our main contributions are summarized as follows:
\begin{itemize}
\item We propose a Learnable Spatial Shift module~(LSS) that significantly expands the receptive field by applying channel-wise spatial shifts, breaking the trade-off between receptive field and computation and storage cost.
\item We adopt and improve the dual-branch design by introducing a radically asymmetric architecture that addresses inherent inefficiencies in prior works and reallocates computational resources for better performance.
\item We design the Error-bounded Adaptive Sampling (EAS) algorithm that automatically determines the optimal sampling stride for LUTs, achieving substantial storage reduction with minimal impact on performance and latency.
\end{itemize}

%% file: sec/2_relatedwork.tex
\section{Related Work}
\label{sec:formatting}

\subsection{LUT-Based Image Restoration}

Look-Up Table (LUT)–based models bypass expensive floating-point operations by precomputing input-output mappings, making them well-suited for deployment on resource-constrained edge devices.
Early work such as SR-LUT~\cite{srlut} constructs spatial-wise LUTs by caching local low-resolution patches and their corresponding high-resolution outputs, effectively transforming a pre-trained network into a computation-efficient LUT. However, these methods suffer from the storage explosion problem, as the required storage size of LUT grows exponentially as the number of indexing entries increases. Specifically, the storage size grows as $B^N$, where $B$ and $N$ denote the data range and the number of index entries, respectively. To address this issue, two lines of work have been extensively studied.

\vspace{-15pt}
\paragraph{Reduce Number of Indexing Entries.} 
By decomposing a single large LUT into multiple compact LUTs with complementary indexing patterns, methods such as MuLUT~\cite{mulut} and HKLUT~\cite{hklut} significantly reduce the number of index entries, thereby mitigating storage overhead.
This idea is further developed in RCLUT~\cite{rclut} and TinyLUT~\cite{tinylut}, which are entirely built with 1-dimensional LUTs with one input entity. They reduce the storage size of an n-dimensional LUT from $B^N$ to $B\times N$, enabling efficient LUT inference. However, these methods enhance performance primarily by cascading multiple LUTs, which leads to a linear increase in overall storage and runtime cost. In comparison, our work proposes a lightweight spatial shift module to improve performance with minimal cost.
\vspace{-15pt}
\paragraph{Reduce Data Range of Indexing Entries.}
Early work, such as SRLUT~\cite{srlut} and MuLUT~\cite{mulut}, reduces the input data range by quantizing the 8-bit input data into a lower-bit range and adopts complex interpolation algorithms to compensate for quantization loss. 
However, the interpolation introduces additional multiplications and comparison operations during inference.
SPLUT~\cite{splut} explores bit-level input decomposition and introduces a symmetric dual-branch structure that processes significant and insignificant bits in parallel, eliminating interpolation and accelerating inference.  
Building on this idea, HKLUT~\cite{hklut} demonstrates that the two branches exhibit different effective receptive fields and adopts an asymmetric design to reduce redundancy and improve efficiency. In this work, we further investigate the asymmetric property in the dual-branch structure and propose a more efficient asymmetric architecture.

\begin{figure*}[t]
\centering
\includegraphics[width=\linewidth]{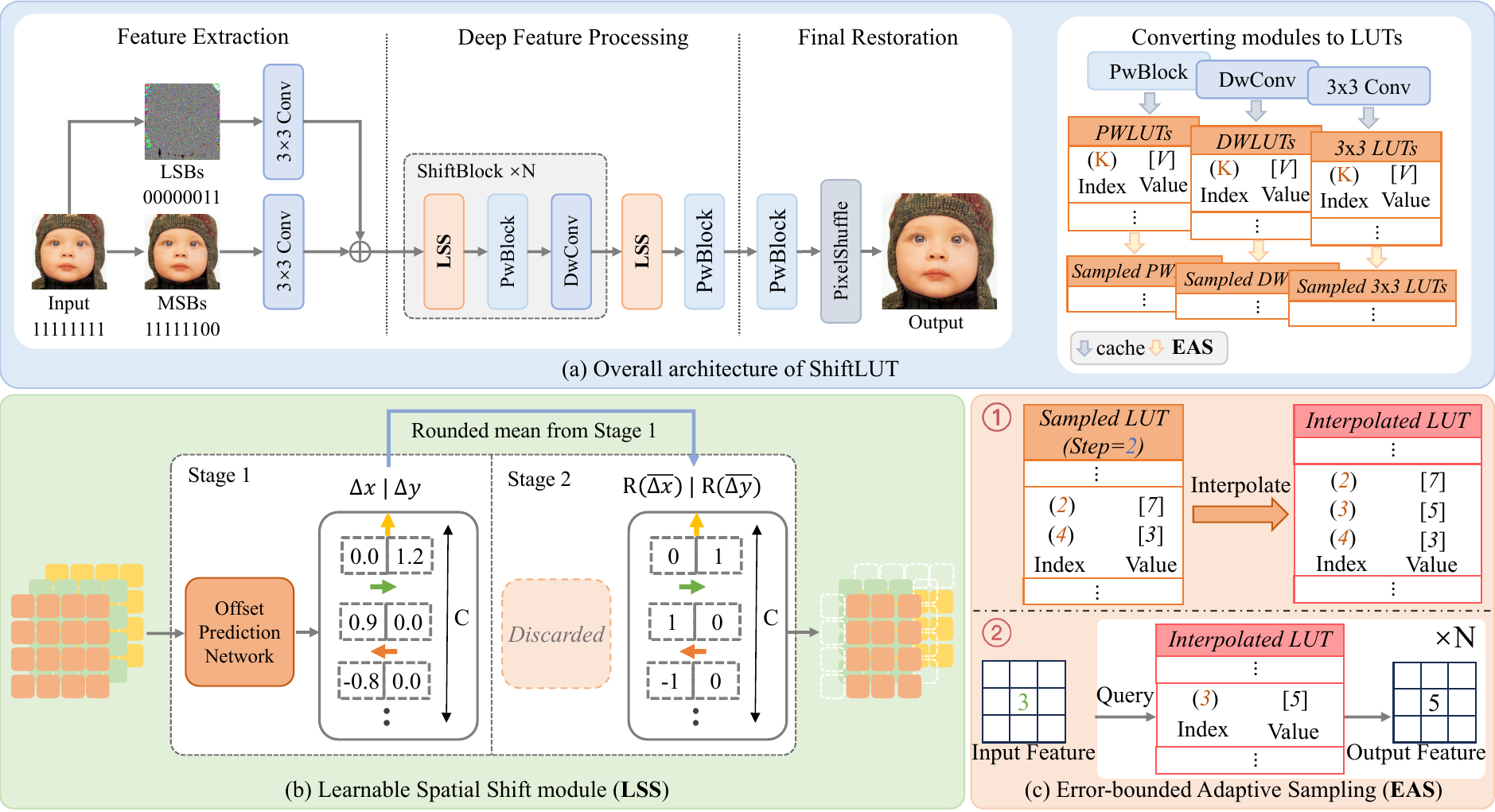} 

\caption{(a) Overall architecture of ShiftLUT. (b) The structure of LSS, which consists of an offset prediction network and a spatial shift operator. In Stage 1, the network predicts floating-point offsets $(\Delta x, \Delta y)$, which are applied via bilinear interpolation. In Stage 2, the offsets are replaced with integer-valued approximations, computed by rounding the average offset from Stage 1. (c) Illustration of the EAS inference pipeline with an example using two sampling steps. EAS precomputes and caches interpolated LUT outputs into a reusable buffer, replacing per-pixel interpolation with a single query operation for faster inference.}
\label{fig:main_arch}
\vspace{-1.0em}
\end{figure*}

\subsection{Feature Shifting}
Feature shifting operates by displacing each channel of the input tensor in a different spatial direction to enable richer feature aggregation. ShiftNet~\cite{wu2018shift} first integrates feature shifting with a 1×1 pointwise convolution as an efficient alternative to standard spatial convolution, and this concept has been further extended to various high-level vision tasks, such as Temporal Shift Module~\cite{lin2019tsm} and AddressNet~\cite{he2019addressnet}.

While the aforementioned studies focus primarily on high-level vision tasks, several works~\cite{pcs, li2023simple, eclut} have explored the integration of shifting operations into low-level vision problems. PCS~\cite{pcs} simplifies the generic feature shifting by imposing task-specific constraints to better suit the SISR task. Group ShiftNet~\cite{li2023simple} expands the effective receptive field by introducing a grouped spatial shift module for VSR. As for LUT-based methods, ECLUT~\cite{eclut} enlarges the receptive field by predicting multiple outputs at different spatial locations, thereby extending the coverage range at the output stage. Although these methods effectively amplify the receptive field with almost negligible computational cost through shifting operations, their shifting offsets remain predefined and fixed. In contrast, our approach introduces Learnable Spatial Shift (LSS), which employs learnable offsets to further enhance overall performance.

%% file: sec/3_method.tex

\section{Method}

\subsection{Overview}

In this work, we propose ShiftLUT, a novel dual-branch network architecture for efficient image restoration. Inheriting the design philosophy of SPLUT~\cite{splut}, our method separates the input features into two branches: one branch processes the most significant six bits (MSB), while the other branch handles the least significant two bits (LSB). As illustrated in \cref{fig:main_arch}(a), our framework adopts a three-stage design:


\vspace{-10pt}
\paragraph{Feature Extraction.}
The input image is first split into its MSB and LSB components. Each component is independently processed by a standard 3×3 convolutional layer to extract shallow features, which are subsequently fused through element-wise addition.

\vspace{-10pt}
\paragraph{Deep Feature Processing.}
The fused shallow features are then fed into a deep processing backbone constructed by stacking multiple Shift-Blocks. Each Shift-Block is sequentially composed of a Learnable Spatial Shift module~(LSS), a Pointwise Block (PwBlock), and a $3\times 3$ depthwise convolution (DwConv). The LSS is introduced to expand the receptive field and the designs of the DwConv and PwBlock 
are adopted from TinyLUT~\cite{tinylut}. To accommodate different deployment scenarios, we design three model variants: ShiftLUT-S, ShiftLUT-M, and ShiftLUT-L, using 0, 1, and 7 stacked Shift-Blocks, respectively.

\vspace{-10pt}
\paragraph{Final Restoration.}
After passing through the stack of shift blocks, the resulting deep features are processed by a final PwBlock for channel refinement. A PixelShuffle layer upscales the features to produce the final result.

Following common practice in LUT-based methods, all convolutional layers are converted into memory-efficient 1D LUTs using the separable mapping strategy (SMS) from TinyLUT~\cite{tinylut}, and the rotation ensemble trick~\cite{srlut} is also adopted to expand the receptive field during inference.
Specifically, $3{\times}3$ standard convolutions are converted into $3{\times}3$ LUTs, the DWConvs become DWLUTs, and the PwBlocks are transformed into PWLUTs.
These strategies ensure that the proposed architecture achieves minimal computational overhead while preserving restoration quality. 


\subsection{Learnable Spatial Shift}
\label{sec:shift}
\paragraph{Design of LSS}
Previous LUT-based methods~\cite{spflut, eclut} have demonstrated that enlarging the receptive field significantly improves performance.
Typical approaches to achieve this include stacking multiple LUT layers in a cascade manner~\cite{splut} or incorporating large kernel convolution and decomposing it with several compact patterns~\cite{mulut}. However, these methods inevitably increase storage and computational costs, conflicting with the lightweight design goals of LUT-based architectures.

Inspired by the works from feature shifting~\cite{wu2018shift,pcs}, which enlarges the receptive field via a lightweight shift operation using fixed offsets on feature maps, we explore a learnable variant tailored for LUT-based methods. Specifically, we propose Learnable Spatial Shift module~(LSS)---an effective solution for expanding receptive fields. 


As illustrated in \cref{fig:main_arch}(b), LSS employs an \emph{Offset Prediction Network}, denoted as $\mathcal{O}(\cdot)$. We implement this network as a lightweight module of several convolutions and an MLP head. This lightweight design is justified by our empirical results, which show that adopting a more complex network yields no performance gain. The network takes an input feature map $\mathbf{F} \in \mathbb{R}^{C \times H \times W}$ and generates a unique pair of displacement values for each channel:
\begin{equation}
\{(\Delta x_c, \Delta y_c)\}_{c=1}^C = \mathcal{O}(\mathbf{F}),
\end{equation}
each channel is then spatially shifted by the offsets:
\begin{equation}
\mathbf{F}'_c(x, y) = \mathbf{F}_c(x - \Delta x_c, \; y - \Delta y_c),
\end{equation}
\begin{figure}[t]
\centering
\includegraphics[width=0.46\textwidth]{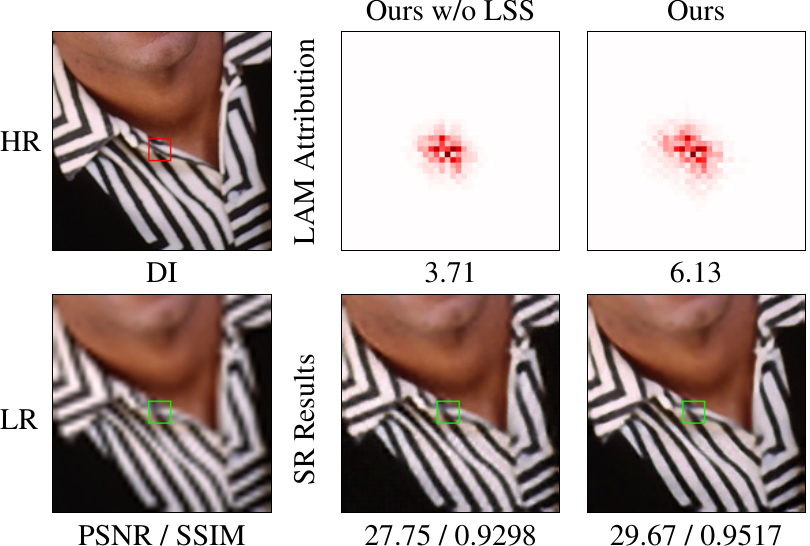} 
\caption{
Local Attribution Map (LAM) visualization for a \(16 \times 16\) output patch.  
A larger DI indicates that a wider range of pixels contributes to the output result.
Our method with LSS shows larger DI and better performance than the variant without LSS.
}
\label{fig:shift}
\end{figure}
each spatial location \((x, y)\) in the shifted feature  \(\mathbf{F}'\) gathers information from a set of \(C\) channel-specific positions in the original input \(\mathbf{F}\).
Afterwards, the subsequent PwBlock fuses these spatially diverse features 
to produce the output feature map \(\mathbf{F}_{\mathrm{out}}\).
As a result, compared to using the PwBlock alone, which would have a 1x1 receptive field, each feature vector \(\mathbf{F}_{\mathrm{out}}(x,y)\) now has effectively ``seen'' a larger, contextually relevant area of the input feature map \(\mathbf{F}\), thus expanding the receptive field.
We empirically validate this receptive field expansion in \cref{fig:shift}, where the Local Attribution Maps (LAM)~\cite{lam} of our model with LSS demonstrate a significantly broader spatial attribution compared to the baseline model without LSS.

\vspace{-10pt}
\paragraph{Two-stage training strategy}
However, deploying the complete LSS at inference time poses two drawbacks. First, the Offset Predict Network and the interpolation step introduce significant computational overhead. Second, directly converting the Offset Predict Network into a LUT is impractical due to an increase in storage. To resolve these issues, we adopt a two-stage training strategy as illustrated in \cref{fig:main_arch}(b). In the first stage, the Offset Predict Network is trained jointly with the rest of the model. In the second stage, the network is removed, and the learned continuous offsets are replaced with fixed integer shifts, computed by rounding the mean offset values collected during training. This allows us to eliminate interpolation at inference time, yielding a more efficient and hardware-friendly design.

To validate our two-stage training strategy, we analyze the variance of predicted offsets across diverse input samples at the end of the first training stage. 
The experiment shows that across all LSS layers, both \(\Delta x\) and \(\Delta y\) consistently exhibit low variance across channels, mostly under $10^{-3}$. 
This indicates that the offsets are stable and largely invariant to input content. Rather than dynamic, content-adaptive shifts, LSS converges to fixed, channel-specific spatial sampling patterns. Thus, the learnable module discovers an optimal static shift configuration. These findings support our training strategy and justify replacing learned continuous offsets with fixed integer shifts during inference.

\subsection{Asymmetric Dual-Branch Architecture}
\label{sec:asy}

The dual-branch structure
was initially proposed by SPLUT~\cite{splut} 
to avoid the computationally expensive interpolation step required by other LUT methods. Its parallel framework applies identical computational complexity to both branches.
The MSB stream represents the image's low-frequency structural content, such as contours and smooth areas, which is spatially dense. Conversely, the LSB stream captures high-frequency details like fine textures and edges, which are known to be inherently sparse in natural images. This intrinsic statistical difference suggests that a symmetric architecture, which applies identical computational complexity to both streams, is suboptimal. While prior work such as HKLUT~\cite{hklut} has justified asymmetry based on differing receptive field requirements, we posit that the core motivation for adopting an asymmetric design lies in the computational redundancy caused by applying complex networks to the sparse LSB stream.

\begin{figure}[t]
\centering
\includegraphics[width=0.45\textwidth]{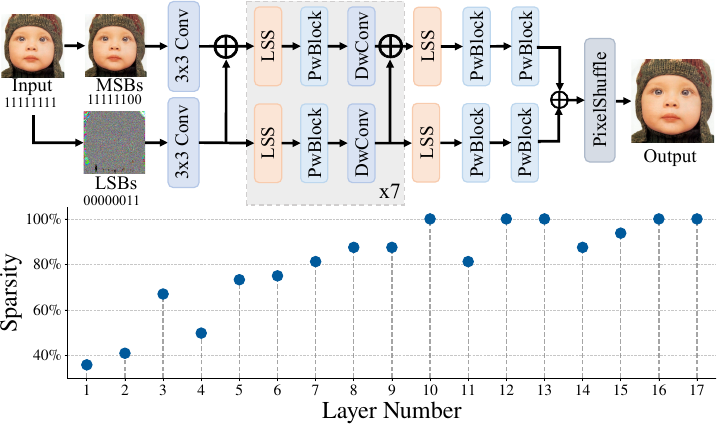} 
\caption{The symmetric network architecture (top) and its corresponding LSB feature sparsity (bottom). The layer-wise analysis shows that feature sparsity in the LSB branch increases significantly with network depth.}
\label{fig:zero}
\vspace{-1.0em}
\end{figure}

To validate this hypothesis, we conducted an experiment by enforcing a deep, symmetric architecture on the LSB branch, identical to that of the MSB counterpart. We then evaluated the resulting activation sparsity throughout the LSB pathway.
As shown in \cref{fig:zero}, the proportion of zero-valued activations in the LSB branch increases with network depth, reaching nearly 100\% in the deeper layers. This result shows that applying a deep, complex network to the LSB stream is highly inefficient, as the vast majority of computations are performed on zero values.



Based on this observation, we propose a novel asymmetric architecture. As illustrated in \cref{fig:main_arch}(a), the LSB branch is simplified to its most essential form: a single $3{\times}3$ convolutional layer. Its output is then directly added to the features of the MSB branch for fusion. This design strategically reallocates computational resources from the LSB to the more information-dense MSB branch, thus eliminating redundancy while maintaining comparable performance. Our experiment shows that the proposed asymmetric design achieves an average PSNR/SSIM of 28.19/0.8014 across five standard SR benchmarks, nearly identical to the 28.19/0.8016 from the symmetric baseline, while reducing inference latency from 164ms to 84ms.





\subsection{Error-bounded Adaptive Sampling}
\label{sec:eas}

The performance of LUT-based models is fundamentally constrained by the size of the lookup table.
To address this limitation, previous studies~\cite{srlut, mulut, spflut} adopt sampling-based strategies that downsample inputs into a lower-bit range and use advanced interpolation to compensate for the sampling loss.
Despite their effectiveness, these methods encounter two major limitations.
First, they rely on a manually predefined sampling stride shared across all LUTs, which is suboptimal since different LUTs contribute unequally to the results.
Second, the complex interpolation algorithms substantially slow down inference.

In this paper, we propose an \textbf{Error-bounded Adaptive Sampling (EAS)} method to overcome these limitations.
EAS consists of two stages (pseudo-code provided in Alg.1 of the supplementary material): (1) Offline Optimization, which automatically determines the optimal sampling stride for each individual LUT, and (2) Online Inference, which utilizes a shared buffer to eliminate interpolation overhead.

Specifically, during the offline stage, the optimal sampling stride is selected from a candidate set
\(
\mathcal{S} = \{2^k \mid k = 0, 1, \dots, K_{\max}\},
\)
where $K_{\max}$ prevents the sampling stride from exceeding the input range. The candidate set, composed of powers of two, is hardware-friendly since modular and division operations with powers of two can be implemented as simple bit shifts.

\begin{table*}[t]
\centering
\renewcommand{\arraystretch}{1.0}
{\fontsize{9pt}{11pt}\selectfont
\setlength{\tabcolsep}{0.7mm}
\begin{tabular}{|l|ccc|cccccc|}
\hline
Method & Storage Size & Runtime & RF & Set5 & Set14 & BSDS100 & Urban100 & Manga109 & Average \\
\hline \hline
FSRCNN & 48KB & 350ms & $17 \times 17$ & 30.71/0.8656 & 27.60/0.7543 & 26.96/0.7129 & 24.61/0.7263 & 27.90/0.8610 & 27.56/0.7840 \\
VDSR & 2660KB & 13681ms & $41 \times 41$ & 31.35/0.8830 & 28.02/0.7680 & 27.29/0.7260 & 25.18/0.7540 & 28.50/0.8812 & 28.07/0.8024 \\
QuickSRNet-S & 33.3KB & 128ms & $9 \times 9$ & 30.91/0.8746 & 27.85/0.7627 & 27.06/0.7183 & 24.76/0.7373 & 28.18/0.8730 & 27.75/0.7932\\
\hline \hline
SPLUT-L & 18432KB & 265ms & $5 \times 5$ & 30.52/0.8630 & 27.54/0.7520 & 26.87/0.7090 & 24.46/0.7191 & 27.70/0.8581 & 27.42/0.7802 \\
MuLUT & 4159KB & 242ms & $9 \times 9$ & 30.60/0.8653 & 27.60/0.7541 & 26.86/0.7110 & 24.46/0.7194 & 27.90/0.8633 & 27.48/0.7826 \\
MuLUT+AutoLUT & 4165KB & 244ms & $17 \times 17$ & 30.85/0.8699 & 27.77/0.7584 & 26.96/0.7144 & 24.60/0.7257 & 28.27/0.8706 & 27.69/0.7878 \\
RCLUT & 1549KB & 232ms & $27 \times 27$ & 30.72/0.8677 & 27.67/0.7577 & 26.95/0.7154 & 24.57/0.7253 & 28.05/0.8655 & 27.59/0.7863 \\
SPF-LUT+DFC & 2066KB & - & $21 \times 21$ & 31.05/0.8755 & 27.88/\textcolor{blue}{\textbf{0.7632}} & 27.08/\textcolor{blue}{\textbf{0.7190}} & 24.81/0.7357 & 28.58/0.8779 & 27.88/0.7943 \\
ECLUT & 9216KB & 41ms & $21 \times 21$ & 29.91/0.8461 & 27.14/0.7419 & 26.61/0.7019 & 23.98/0.6977 & 26.96/0.8362 & 26.92/0.7647 \\
TinyLUT-S & \textcolor{blue}{\textbf{37KB}} & \textcolor{blue}{\textbf{29ms}} & $5 \times 5$ & 30.22/0.8535 & 27.33/0.7450 & 26.71/0.7042 & 24.19/0.7066 & 27.21/0.8458 & 27.13/0.7710 \\
TinyLUT-F & 171KB & 146ms & $\textcolor{blue}{\mathbf{33 \times 33}}$ & \textcolor{blue}{\textbf{31.18}}/\textcolor{blue}{\textbf{0.8771}} & \textcolor{blue}{\textbf{28.01}}/0.7630 & \textcolor{blue}{\textbf{27.13}}/0.7184 & \textcolor{blue}{\textbf{24.92}}/\textcolor{blue}{\textbf{0.7397}} & \textcolor{blue}{\textbf{28.83}}/\textcolor{blue}{\textbf{0.8798}} & \textcolor{blue}{\textbf{28.01}}/\textcolor{blue}{\textbf{0.7956}} \\ 
\hline \hline
ShiftLUT-S~(\textbf{ours}) & \textcolor{red}{\textbf{24KB}} & \textcolor{red}{\textbf{22ms}} & $9 \times 9$ & 30.50/0.8609 & 27.54/0.7523 & 26.87/0.7103 & 24.39/0.7169 & 27.65/0.8563 & 27.39/0.7793 \\
ShiftLUT-M~(\textbf{ours}) & 38KB & 31ms & $17 \times 17$ & 30.77/0.8683 & 27.74/0.7582 & 26.98/0.7149 & 24.62/0.7272 & 28.18/0.8682 & 27.66/0.7874 \\
ShiftLUT-L~(\textbf{ours}) & 104KB & 84ms & $\textcolor{red}{\mathbf{65 \times 65}}$ & \textcolor{red}{\textbf{31.33}}/\textcolor{red}{\textbf{0.8798}} & \textcolor{red}{\textbf{28.11}}/\textcolor{red}{\textbf{0.7686}} & \textcolor{red}{\textbf{27.21}}/\textcolor{red}{\textbf{0.7230}} & \textcolor{red}{\textbf{25.12}}/\textcolor{red}{\textbf{0.7487}} & \textcolor{red}{\textbf{29.16}}/\textcolor{red}{\textbf{0.8870}} & \textcolor{red}{\textbf{28.19}}/\textcolor{red}{\textbf{0.8014}} \\
\hline
\end{tabular}
}
\caption{Quantitative comparisons on 5 standard SISR test sets for an upscaling factor of 4. The best and second-best results of each metric are highlighted in \textcolor{red}{\textbf{red}} and \textcolor{blue}{\textbf{blue}}, respectively (the highlighting is restricted to LUT-based methods to emphasize fair comparison).}
\label{tab:sr}
\end{table*}

Within this candidate set, the optimal stride corresponds to the maximum $s$ whose interpolation error(defined in~\cref{eq:eas_loss}) remains below the predefined tolerance~$\varepsilon$
\begin{equation}
\begin{aligned}
& \max_{s \in \mathcal{S}} \; s 
\quad \text{s.t.} \quad \mathrm{Error}(s) < \varepsilon,\\[3pt]
\mathrm{Error}(s)
&= \tfrac{s}{s-1} \cdot
\mathbb{E}_{i \sim \mathcal{I}}
\!\left[
\big|
\mathrm{Query}_{s}(i) - \text{LUT}[i]
\big|
\right].
\end{aligned}
\label{eq:eas_loss}
\end{equation}
where $\mathcal{I}$ denotes the index set of the original LUT, and $\mathrm{Query}_s(i)$ represents the linearly interpolated LUT output at index~$i$ under stride~$s$.
The expectation term $\mathbb{E}[\cdot]$ is weighted by
$\tfrac{s}{s-1}$, which serves as a penalty weight derived from the storage saving rate. 



During the online inference stage, we accelerate the inference of EAS by caching intermediate interpolation results.
In conventional sampling-based methods, inference is slowed down by repeated interpolation computations performed for every pixel. 
As illustrated in ~\cref{fig:main_arch}(c), to eliminate this redundancy, we precompute the interpolated LUT outputs $\mathrm{Query}_{s}(i)$ once at inference time and cache them in a shared reusable buffer for each LUT. 
Each pixel thus directly queries these cached values instead of performing repeated interpolations.
This caching strategy introduces negligible memory overhead,
since the buffer size is determined solely by the input bit-width (e.g., for a 6-bit input range, the buffer contains only $2^6=64$ entries, requiring just $64$ Bytes).

\begin{figure*}[t]
    \centering
    \includegraphics[width=1.0\linewidth]{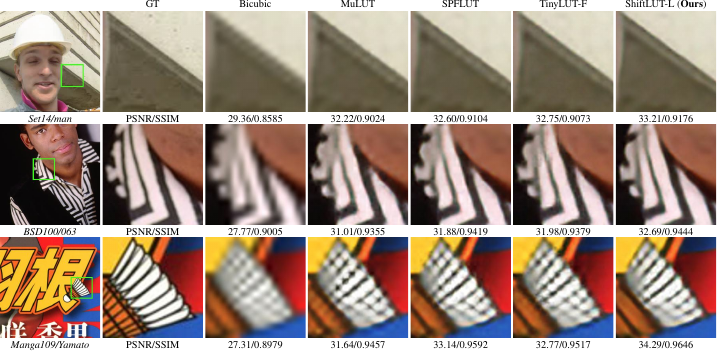}
    \caption{Qualitative comparison for $4\times$ super-resolution on different images.}
    \label{fig:sr}
\end{figure*}

%% file: sec/4_experiments.tex
\section{Experiments}

\subsection{Implementation Details}

We evaluate our method on three image restoration tasks, including classic image super-resolution, image denoising and image deblocking. And for all tasks, we use DIV2K~\cite{div2k} as our training dataset. The model is trained for 200{,}000 iterations using the Adam optimizer~\cite{adam} with parameters \( \beta_1 = 0.9 \), \( \beta_2 = 0.999 \), and an initial learning rate of \( 5 \times 10^{-3} \). The learning rate is gradually decreased following a cosine annealing schedule~\cite{cosineannealing}. During training, input patches of size \( 48 \times 48 \) are randomly cropped from low-resolution images and augmented with random flipping and rotation. The batch size is set to 32. While for inference, the tolerance~$\varepsilon$ in EAS is set to 0.4.

For the denoising and deblocking tasks, we modify the network by removing the PixelShuffle layer and setting the output channels of the final PwLUT module to 1. 

\begin{table}[t]
\centering
\renewcommand{\arraystretch}{1.1}
\setlength{\tabcolsep}{1mm}
\begin{tabular}{|llccc|}
\hline
& Method & Storage Size & Set12 & BSD68 \\ \hline
\multirow{4}{*}{LUT} & MuLUT & 489KB & 31.50 & 30.63 \\ 
& SPFLUT+DFC & 596KB & 32.01 & 31.09 \\ 
& TinyLUT-F & 187KB & 32.22 & 31.20 \\ 
& ShiftLUT-L~(\textbf{ours}) & \textcolor{red}{\textbf{60KB}} & \textcolor{red}{\textbf{32.43}} & \textcolor{red}{\textbf{31.32}} \\ \hline
\multirow{1}{*}{DNN} & DnCNN & 2220KB & 32.86 & 31.73 \\ \hline
\end{tabular}
\caption{The comparison for grayscale image denoising at a noise level of 15 on standard benchmark datasets. The best of each metric is highlighted in \textcolor{red}{\textbf{Red}}.}
\label{tab:denoising}
\end{table}

\begin{figure}[t]
    \centering
    \includegraphics[width=1.0\linewidth]{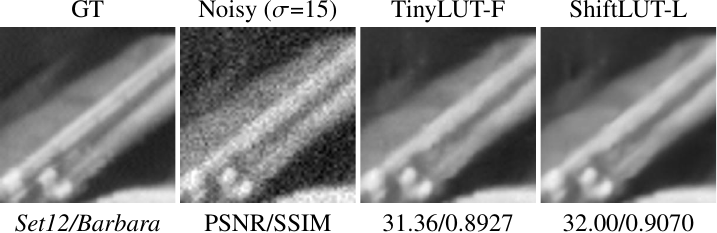}
    \caption{Qualitative comparison for image denoising}
    \label{fig:denoising}
    \vspace{-1.0em}
\end{figure}

\subsection{Evaluation on Image Super Resolution}

We evaluate the performance of our model on five standard benchmark datasets for \( \times 4 \) super-resolution: Set5, Set14, BSDS100~\cite{bsds100}, Urban100~\cite{urban100}, and Manga109~\cite{manga109}. The low-resolution images are generated by bicubic downsampling from their corresponding high-resolution ground truths. For quantitative assessment, we compute PSNR and SSIM on the Y channel of the YCbCr color space, following common practice in the SISR literature~\cite{edsr,rcan}. Besides, we measured and reported the runtime on Xiaomi 11 smartphones with a Qualcomm Snapdragon 888 CPU, using an on-device Android application similar to SPLUT, where all methods were evaluated at an input resolution of 320 $\times$ 180 to ensure fair comparison. 

\cref{tab:sr} compares our method against several state-of-the-art LUT-based and DNN-based methods, 
including SPLUT-L~\cite{splut}, MuLUT~\cite{mulut}, RCLUT~\cite{rclut}, SPF-LUT+DFC~\cite{spflut}, ECLUT~\cite{eclut}, TinyLUT-F~\cite{tinylut} and AutoLUT~\cite{autolut}, as well as DNN-based approaches such as FSRCNN~\cite{fsrcnn}, VDSR~\cite{vdsr} and QuickSRNet-S~\cite{berger2023quicksrnet}.
Our ShiftLUT framework produces a family of models that establish a new Pareto frontier for LUT-based single image super-resolution.
Our largest model, ShiftLUT-L, sets a new state-of-the-art among LUT-based methods, outperforming the previous best, TinyLUT-F, across all standard benchmarks. On Manga109, it improves PSNR from 28.83~dB to 29.16~dB, while reducing the LUT size from 171~KB to 104~KB and decreasing runtime from 146~ms to 84~ms.
On the compact end, ShiftLUT-S achieves the smallest storage and fastest runtime, making it the most efficient overall.
In between, ShiftLUT-M offers a compelling trade-off, outperforming the well-established DNN-based FSRCNN in average PSNR while being over 11$\times$ faster and requiring just 42~KB of storage.
These results demonstrate that ShiftLUT provides a versatile suite of models that advance the trade-off between fidelity, storage, and inference speed.

We provide the visual comparison in \cref{fig:sr}. For the images from Set14, BSDS100 and Manga109, our results show clearer edges and finer structural details, such as restoring intricate line work in manga images or reconstructing complex textures and geometric patterns in natural scenes. 
In contrast, the other approaches, such as MuLUT and SPFLUT, all suffer from blurring and loss of detail. These examples together highlight our method’s strong ability to preserve sharpness and structure, resulting in more natural and visually pleasing reconstructions.

\begin{table}[t]
\centering
\renewcommand{\arraystretch}{1.1}
\setlength{\tabcolsep}{1mm}
\begin{tabular}{|llccc|}
\hline
 & Method & Storage Size & Classic5 & LIVE1 \\ \hline
\multirow{4}{*}{LUT}  & MuLUT & 489KB & 28.29 & 28.39 \\ 
& SPFLUT+DFC & 596KB & 28.62 & 28.61 \\ 
& TinyLUT-F & 187KB & 28.74 & 28.67 \\ 
& ShiftLUT-L~(\textbf{ours}) & \textcolor{red}{\textbf{60KB}} & \textcolor{red}{\textbf{29.12}} & \textcolor{red}{\textbf{28.96}} \\ \hline
\multirow{1}{*}{DNN} & ARCNN & 415KB & 28.76 & 28.77  \\ \hline
\end{tabular}
\caption{The comparison for image deblocking under a quality factor of 10 on standard benchmark datasets. The best of each metric is highlighted in \textcolor{red}{\textbf{Red}}.}
\label{tab:deblocking}
\end{table}

\begin{figure}[t]
    \centering
    \includegraphics[width=1.0\linewidth]{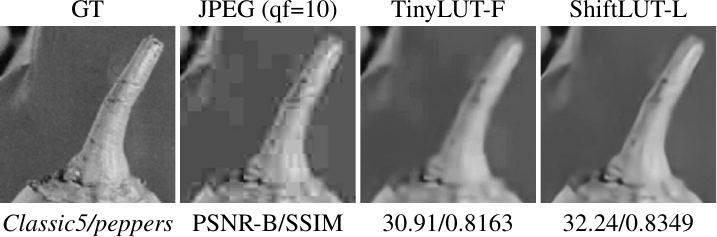}
    \caption{Qualitative comparison for image deblocking}
    \label{fig:deblocking}
    \vspace{-1.0em}
\end{figure}

\subsection{Evaluation on Image Denoising}

We conduct evaluations of LUT-based models for grayscale image denoising using the Set12 ~\cite{set12} and BSD68 ~\cite{bsd68} benchmark datasets, with the noise level set to 15. The noisy images are generated by adding Gaussian white noise.
The quantitative results of image denoising is shown in \cref{tab:denoising}.
As observed, our method achieves better denoising performance with a smaller storage size. Specifically, it improves the PSNR from 32.22 dB to 32.43 dB on Set12 and from 31.20 dB to 31.32 dB on BSD68, compared to the previous best LUT-based model.

We provide qualitative comparisons in \cref{fig:denoising} to further demonstrate the effectiveness of our method. 
TinyLUT-F exhibits residual artifacts and blurring, especially in regions with fine structures. 
In contrast, our method not only removes the noise more thoroughly but also better preserves edge sharpness and subtle textures, resulting in a denoised image closer to the ground truth.
These visual results are consistent with the quantitative improvements reported in \cref{tab:denoising}, confirming that our approach achieves superior noise removal while maintaining image details.

\subsection{Evaluation on Image Deblocking}

For the image deblocking task, we evaluate all methods on the Classic5 and LIVE1 ~\cite{live1} datasets, using a JPEG compression quality factor of 10 to simulate severe compression artifacts. The PSNR-B metric ~\cite{yim2010quality} is employed to quantitatively assess the reduction of blocking artifacts.

\cref{tab:deblocking} presents the quantitative results. Our method achieves the highest PSNR-B among all compared approaches, including the CNN-based ARCNN. Specifically, it improves the PSNR-B from 28.74 dB to 29.12 dB on Classic5 and from 28.67 dB to 28.96 dB on LIVE1, compared to TinyLUT-F.
As shown in \cref{fig:deblocking}, our method more effectively suppresses blocking artifacts while preserving fine image details. Compared to TinyLUT-F, it produces smoother transitions in homogeneous regions and sharper edges, leading to visually more pleasing results.





\subsection{Ablation Studies}
\label{sec:ablation}



\begin{figure}[t]
    \centering
    \includegraphics[width=1.0\linewidth]{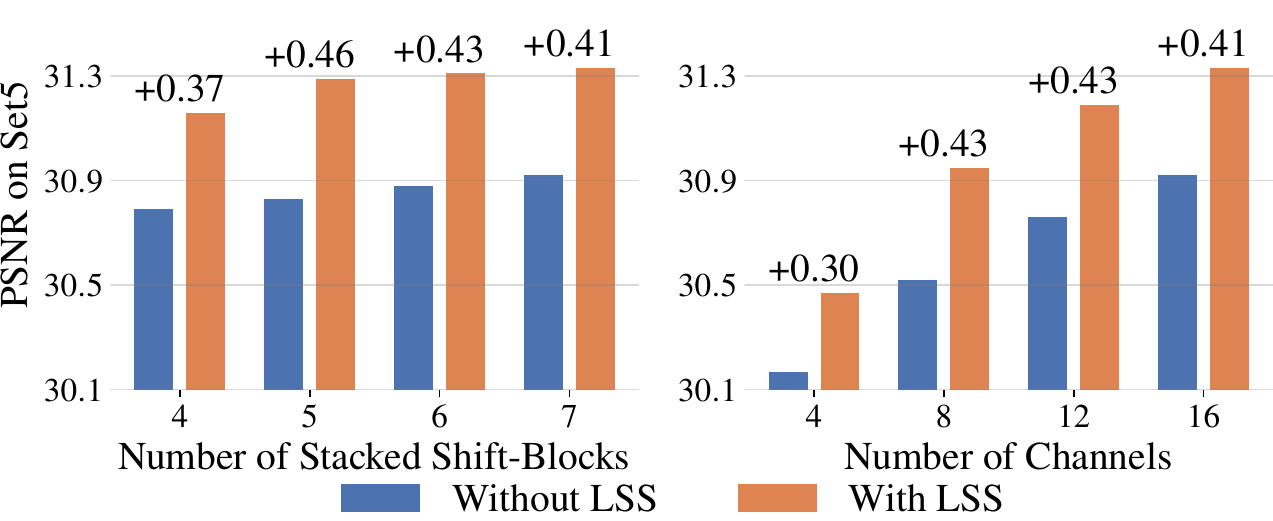}
    \caption{
    PSNR comparison on the Set5 under different network configurations. 
    The left and right figures show results obtained by varying the number of stacked ShiftBlocks and channels, respectively. 
    Each bar group compares models with and without LSS.
    }
    \label{fig:ablation_config}
\end{figure}

\paragraph{Ablation on LSS under Different Configurations.}
To evaluate the effectiveness of LSS, we conduct experiments by comparing models with and without LSS under different network configurations
by varying the number of ShiftBlock stacks and channel dimensions within the network. As shown in \cref{fig:ablation_config}, incorporating LSS consistently yields PSNR improvements exceeding 0.30\,dB on the Set5 benchmark across all tested configurations, illustrating that LSS serves as a universally effective module, delivering stable and notable gains regardless of the network configuration.

\begin{table}[t]
\centering
\renewcommand\arraystretch{1.1}
\caption{
Compare EAS with other LUT compression methods.
}
{\fontsize{9pt}{11pt}\selectfont 
\setlength{\tabcolsep}{2.5pt}
\begin{tabular}{lccccc}
\hline
Method & Set5 & Set14 & B100 &  LUT Size & Runtime \\
\hline
ShiftLUT & 31.33 & 28.11 & 27.21 & 217KB & 81ms \\
\hline
+DDM & 31.26 &	28.06 &	27.19 & 172KB & 98ms
 \\ \hline
+Uniform(Step=2) & 
31.33 & 28.11 & 27.20  & 112KB & 213ms
\\ \hline
+Uniform(Step=4) & 
31.29 & 28.07 & 27.18 & 58KB & 211ms
\\ \hline
+EAS($\varepsilon=0.4$) & 31.33 & 28.11 & 27.21 & 104KB & 84ms
 \\
\hline
+EAS($\varepsilon=0.8$) & 31.30 & 28.10 & 27.19
 & 54KB & 84ms
 \\
\hline
\end{tabular}
}
\label{tab:ablation_eas} 
\vspace{-1.0em}
\end{table}

\vspace{-10pt}
\paragraph{Ablation on EAS.}
In \cref{tab:ablation_eas}, we compare our proposed EAS with other LUT compression methods on the Set5 benchmark, including the dynamic discretization mechanism (DDM) in TinyLUT~\cite{tinylut} and the uniform sampling strategy in SRLUT~\cite{srlut}. Other approaches, such as the DFC module in SPFLUT~\cite{spflut}, are excluded as they are incompatible with the ShiftLUT framework. As shown in Table \ref{tab:ablation_eas}, our proposed EAS consistently achieves a superior trade-off between accuracy, model size, and runtime.
Specifically, EAS($\varepsilon=0.4$) maintains identical PSNR to the original ShiftLUT while reducing LUT storage by over 50\%.
Even under more aggressive compression (EAS($\varepsilon=0.8$)), the performance degradation remains negligible ($<$ 0.03 dB), whereas Uniform sampling and DDM suffer larger losses.
Moreover, EAS restores the runtime efficiency close to the uncompressed baseline, avoiding the heavy interpolation overhead of Uniform sampling.
These results demonstrate that EAS effectively balances reconstruction fidelity, compression efficiency, and inference speed. 

\begin{figure}[t]
    \centering
    \includegraphics[width=1.0\linewidth]{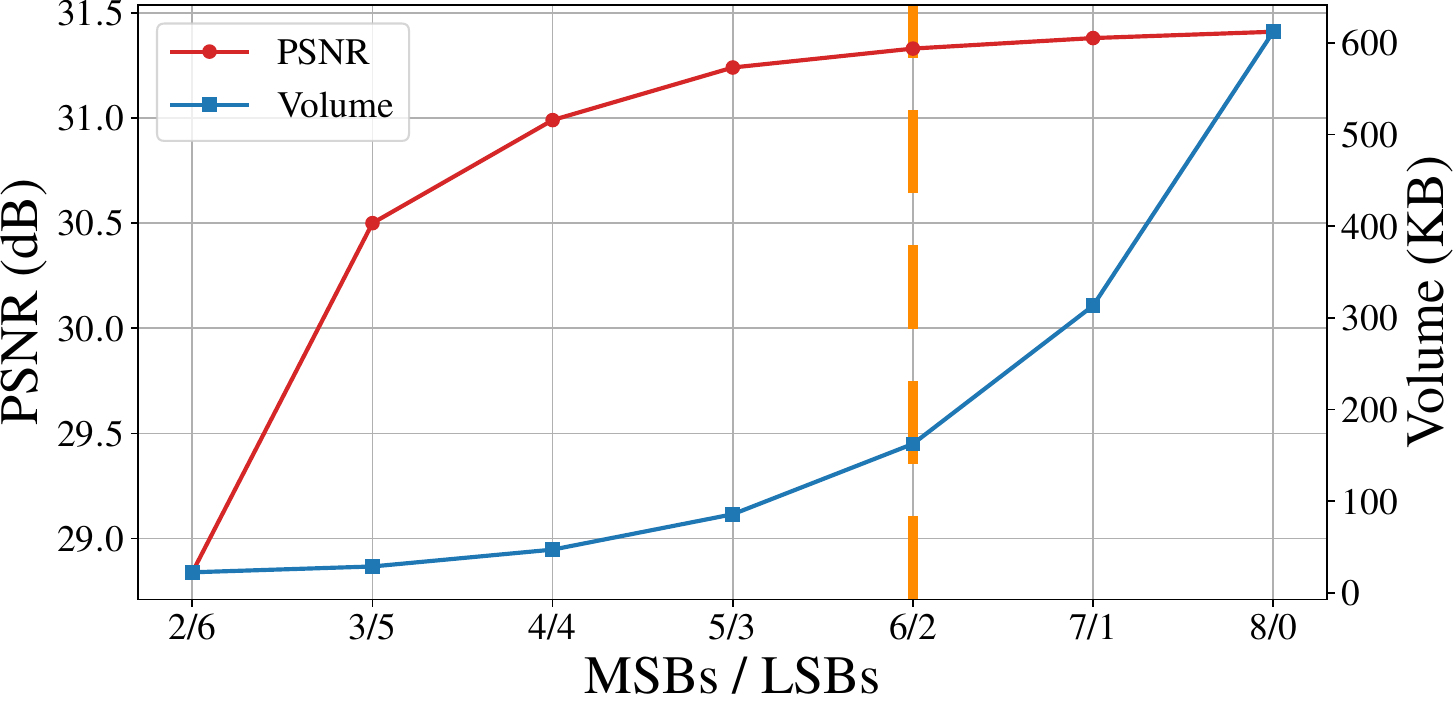}
    \caption{
    Comparison for different MSB/LSB splits.
    }
    \label{fig:dual}
    \vspace{-1.0em}
\end{figure}

\vspace{-10pt}
\paragraph{Ablation on Different MSB/LSB Combination.}
We evaluate the impact of different MSB/LSB bitwidth allocations under a fixed 8-bit activation precision. As shown in \cref{fig:dual}, all possible allocation configurations within the ShiftLUT framework are examined. The results indicate that assigning more bits to the MSB branch generally enhances accuracy but leads to a substantial increase in LUT storage overhead. Among all candidates, the 6/2 configuration achieves a favorable balance between performance and memory consumption, and is adopted as the default.

%% file: sec/5_conclusion.tex
\section{Conclusion}

In this paper, we propose ShiftLUT to effectively expand the receptive field by introducing LSS, a learnable, channel-wise spatial shift module. In addition, we propose an asymmetric dual-branch architecture that assigns more computational resources to important information. Additionally, EAS technique significantly reduces LUT storage size, making the model more practical for resource-constrained devices. Comprehensive experiments show that our method outperforms existing LUT-based approaches.


%% file: sec/X_suppl.tex
\clearpage
\setcounter{page}{1}
\maketitlesupplementary

\setcounter{section}{0}
\renewcommand\thesection{S\arabic{section}}
In this supplementary file, we provide:
\begin{itemize}
    \item Preliminary knowledge for the training, transferring, and testing process of LUT-based methods, as well as the SMS strategy in ~\cref{sec:preliminary}.
    \item An additional ablation study on comparing alternative shifting operations with LSS in ~\cref{sec:shiftop}.
    \item An additional ablation study on integrating LSS into other LUT-based methods in ~\cref{sec:lss}.
    \item More quantitative and qualitative comparisons among LUT methods on image restoration tasks in ~\cref{sec:visual}.
    \item Pseudo code of EAS in ~\cref{sec:code_eas}.
\end{itemize}

\section{Preliminary}
\label{sec:preliminary}


\begin{figure*}[t]
\centering
\includegraphics[width=\linewidth]{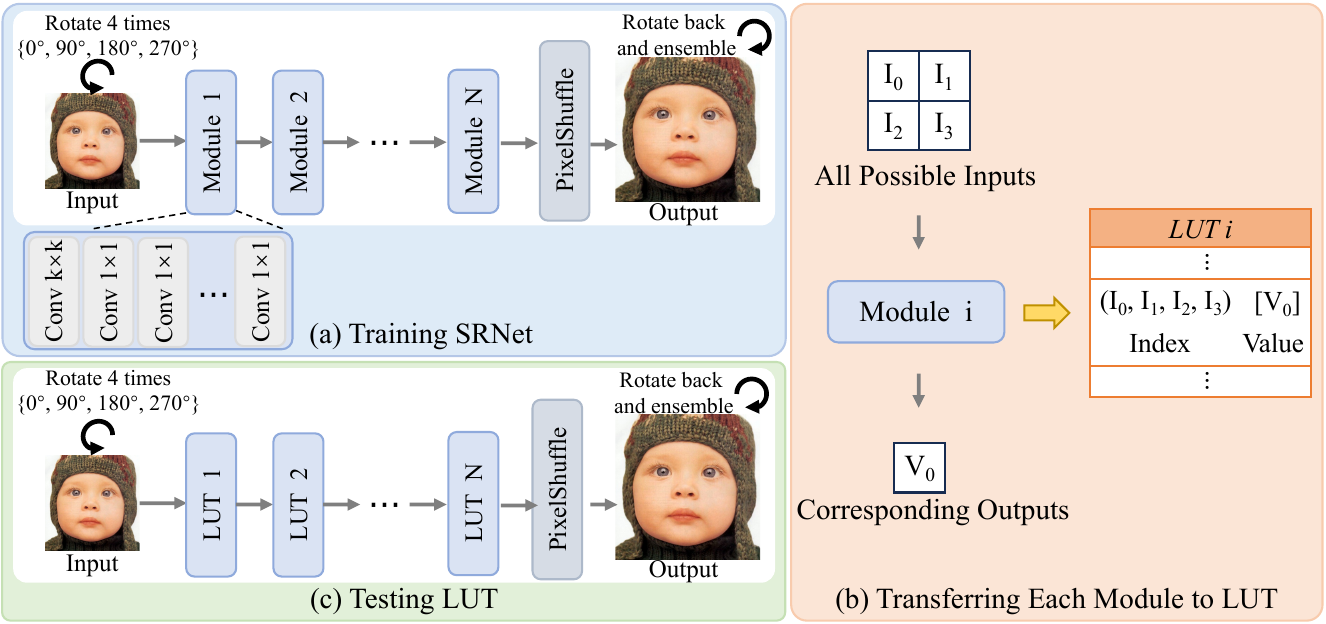} 
\caption{Illustration of training, transferring and testing process for LUT-based methods.}
\label{fig:supp_lut}
\end{figure*}

Lookup Table (LUT) based super-resolution aims to replace expensive pixel-wise computations with efficient table lookup and interpolation. The core idea is to train a lightweight SR model with a constrained receptive field (RF). Once the model is fully trained, enumerates all possible input configurations within this RF and stores the corresponding high-resolution (HR) outputs in a precomputed LUT. During inference, the HR value is obtained exclusively by querying and interpolating LUT entries, yielding extremely fast runtime. As illstrated in ~\cref{fig:supp_lut}, the overall process of LUT-based methods involves three primary stages: (1) training a neural network, (2) transferring its learned mapping into LUTs, and (3) performing inference entirely through LUT queries.


\vspace{-10pt}
\paragraph{Training.} 
Let $x_i \in \mathbb{R}^{k \times k}$ denote the low-resolution (LR) input patch corresponding to a receptive field of size $k \times k$ pixels. A compact CNN $f(\cdot)$ constrained to this RF is trained such that it predicts $r^2$ HR sub-pixels following the pixel-shuffle paradigm:
\begin{equation}
    \hat{y}_i = f(x_i), \qquad \hat{y}_i \in \mathbb{R}^{r^2},
\end{equation}
where $r$ is the upsampling factor. To enlarge the effective context while keeping RF small, the model is trained using $4$ rotational augmentations:
\begin{equation}
    \hat{y}_i = \frac{1}{4} \sum_{j=0}^{3} R_j^{-1}\!\left(f(R_j(x_i))\right),
    \label{eq:rot_ensemble}
\end{equation}
where $R_j$ rotates the patch by $j\times90^\circ$. The trained network is only used once to build the LUT, and it is not required during inference.

\vspace{-10pt}
\paragraph{Transferring.}
After training, enumerates every possible input configuration within the RF space and stores the network output into an $n$-dimensional LUT, where $n=k^2$ is the number of pixels in the receptive field.

Because a full LUT with $2^8$ bins per dimension is prohibitively large for $n\ge3$, previous studies adopt uniform sampling with interval $s$ (e.g., $s=16$ or $24$) to address this issue. Each input pixel value $v\in[0,255]$ is quantized by:
\begin{equation}
    \tilde{v} = \operatorname{clip}\!\left(\left\lfloor \frac{v}{s} \right\rfloor, 0, N-1\right),
\end{equation}
where $N = \frac{256}{s}+1$ is the number of sampling bins. For each sampled tuple $(\tilde{v}_1,\dots,\tilde{v}_n)$ compute:
\begin{equation}
    \mathrm{LUT}[\tilde{v}_1,\dots,\tilde{v}_n] = f(v_1,\dots,v_n),
\end{equation}
where $(v_1,\dots,v_n)$ is the corresponding representative input. This produces an $n$-dimensional array storing $r^2$ HR sub-pixels per entry.

\vspace{-10pt}
\paragraph{Testing.}
At test time, the HR reconstruction is performed solely via memory lookup and multidimensional interpolation. Given an LR patch $(v_1,\dots,v_n)$, it identifies the nearest sampled bins by extracting the most significant bits:
\begin{equation}
    \tilde{v}_i = \left\lfloor \frac{v_i}{s} \right\rfloor,
    \quad
    L_i = v_i \bmod s ,
\end{equation}
where $L_i$ is the residual offset used for interpolation.

For $n$-dimensional LUTs, simplex-based interpolation (triangle in 2D, tetrahedron in 3D, 4-simplex in 4D) is adopted. Let $\{O_0,\dots,O_n\}$ be the LUT vertices of the selected simplex and $\{w_0,\dots,w_n\}$ the corresponding barycentric weights derived from $(L_1,\dots,L_n)$. The final HR output is:
\begin{equation}
    \hat{y} = \frac{1}{s} \sum_{i=0}^{n} w_i O_i.
\end{equation}

This inference pipeline contains no convolution and only involves table lookup and a few integer operations, enabling extremely fast execution on resource-limited devices.




\paragraph{Separable Mapping Strategy (SMS).} A major challenge in converting convolutions to LUTs is the exponential growth of storage with kernel size. For example, a $3\times3$ convolution requires a 9-dimensional LUT, which is impractical.

SMS~\cite{tinylut} addresses this by decomposing an $k\times k$ convolution into $k^2$ independent $1\times1$ mappings, each represented by a simple 1D LUT. This reduces the storage dependency from exponential ($v^{k^2}$) to linear ($s\times k^2$), where $v$ denotes the number of sampled values. The final output is obtained by averaging the results of all sub-LUTs:
\begin{equation}
\hat{F}{out}=\frac{1}{k^{2}}\sum_{i=0}^{n-1}  \sum_{j=0}^{n-1}LUT_{(i,j)}[x_{(i,j)}],
\end{equation}
where $x_{(i,j)}$ is the input feature and $LUT_{(i,j)}$ the corresponding 1D LUT.

\section{Comparison of other shifting operations}
\label{sec:shiftop}

In Section~{3.2}, we introduce LSS with a two-stage training strategy: in the first stage, the offset network predicts floating-point offsets, while in the second stage, the fixed integer-valued offsets are adopted. This quantization process inevitably degrades performance. As shown in \cref{tab:ablation_shift}, there is a 0.017dB performance drop between the two stages. 

A natural question arises: \emph{can LSS directly predict integer-valued offsets to avoid such degradation}? To investigate this, we explore two variants of the LSS module with discrete learnable offsets: the first approach employs Straight-Through Estimation (STE) to round the output offsets from the offset network while maintaining differentiability. The second approach adopts the Gumbel-Softmax method~\cite{jang2016categorical}, which is commonly used in reinforcement learning to model discrete action states. Experimental results in ~\cref{tab:ablation_shift} show that our two-stage training consistently attains the highest reconstruction accuracy across all test sets by a large margin, which also validates the effectiveness of the two-stage training strategy proposed in the main paper.

\begin{table}[h]
\centering
\renewcommand\arraystretch{1.1}
\caption{
Comparison of different shifting operations in ShiftLUT on standard SISR test sets for an upscaling factor of 4.
}
{\fontsize{9pt}{11pt}\selectfont 
\setlength{\tabcolsep}{2.5pt}
\begin{tabular}{lcccc}
\hline
Method & Set5 & Set14 & BSDS100 & Urban100  \\
\hline
ShiftLUT (First Stage) & 31.35 & 28.12 & 27.23 & 25.14
\\
\hline
ShiftLUT (Second Stage) & 31.33 & 28.11 & 27.21 & 25.12 
 \\
\hline
ShiftLUT (STE) & 30.91 & 27.83 & 27.04 & 24.77 
 \\
\hline
ShiftLUT (Gumbel Softmax) & 31.11 & 27.97 & 27.12 & 24.95 
 \\
\hline
\end{tabular}
}
\label{tab:ablation_shift} 
\end{table}

\begin{table*}[t]
\centering
\caption{Quantitative comparison of PSNR on standard benchmark datasets for x4 super-resolution tasks between the original
version of SPLUT and TinyLUT, and the modified version integrating LSS.}
\label{tab:supp_lss}
\begin{tabular}{lcccccc}
\hline
 & \textit{Storage Size} & \textit{Set5} & \textit{Set14} & \textit{BSDS100} & \textit{Urban100} & \textit{Manga109} \\
\hline
SPLUT         & 5632KB & 30.01 & 27.20 & 26.68 & 24.13 & 27.00 \\
\rowcolor{gray!30}
SPLUT + LSS                 & 5632KB &  30.03 & 27.23 & 26.65 &  24.13 & 27.05 \\
TinyLUT         & 171KB & 31.18 & 28.01 & 27.13 & 24.92 & 28.83 \\
\rowcolor{gray!30}
TinyLUT + LSS                 & 171KB & 31.24 & 28.04 & 27.17 & 25.05 & 28.98 \\
\hline
\end{tabular}
\end{table*}

\begin{table*}[h]
\centering
\renewcommand{\arraystretch}{1.1}
\setlength{\tabcolsep}{1mm}
\caption{PSNR-B results of different methods on Classic5 and LIVE1 (quality factor 20, 30, 40).}
\label{tab:supp_deblock}
\begin{tabular}{llcccccc}
\hline
\textbf{PSNR-B} & \textbf{Method} & \multicolumn{3}{c}{\textbf{Classic5}} & \multicolumn{3}{c}{\textbf{LIVE1}} \\
 & & 20 & 30 & 40 & 20 & 30 & 40 \\
\hline
\multirow{4}{*}{LUT}
  & SR-LUT         & 29.54 & 30.80 & 31.71 & 29.74 & 30.99 & 32.00 \\
  & MuLUT          & 30.30 & 31.57 & 32.32 & 30.52 & 31.79 & 32.66 \\
  & SPF-LUT+DFC    & 30.65 & 31.95 & 32.75 & 30.80 & 32.10 & 33.02 \\
  & \textbf{ShiftLUT-F(ours)}       & \textcolor{red}{31.15} & \textcolor{red}{32.33} & \textcolor{red}{33.15} & \textcolor{red}{31.07} & \textcolor{red}{32.31} & \textcolor{red}{33.29} \\
\hline
\multirow{2}{*}{DNN}
  & ARCNN     & 30.59 & 31.98 & 32.79 & 30.79 & 32.38 & 33.14 \\
  & SwinIR   & 31.99 & 33.03 & 33.66 & 31.70 & 33.01 & 33.88 \\
\hline
\end{tabular}
\end{table*}

\section{Generalization of the LSS Module}
\label{sec:lss}

To verify the generalization ability of the proposed LSS module, we integrate LSS into two representative LUT-based networks and report the resulting restoration accuracy for $\times4$ SISR in ~\cref{tab:supp_lss}. 
We choose SPLUT and TinyLUT as the target backbones because both architectures expose multi-channel intermediate features ($c>1$), which is a necessary condition for LSS. 

As shown in ~\cref{tab:supp_lss}, when LSS is appended to SPLUT (feature channels $c=4$), the effect is marginal: Set5 increases by $+0.02$ dB (30.01 $\rightarrow$ 30.03), Set14 by $+0.03$ dB, Manga109 by $+0.05$ dB. 
By contrast, inserting LSS into TinyLUT (feature channels $c=16$) yields consistently larger gains across all datasets: Set5 $+0.06$ dB (31.18 $\rightarrow$ 31.24), Set14 $+0.03$ dB, BSDS100 $+0.04$ dB, Urban100 $+0.13$ dB and Manga109 $+0.15$ dB. 

From these observations we draw two conclusions. First, LSS generalizes across different LUT-based architectures: integrating LSS into both SPLUT and TinyLUT produces non-negative or improved reconstruction performance, demonstrating applicability beyond the original backbone. 
Second, the magnitude of the benefit correlates with the intermediate feature channel dimensionality: networks with more channels (TinyLUT, $c=16$) obtain larger and more consistent PSNR improvements than networks with fewer channels (SPLUT, $c=4$). 
This behaviour is consistent with the design rationale of LSS, which leverages channel-wise spatial diversity into LUTs; hence, LSS is expected to be more effective when richer channel-wise representations are available.

\section{Additional Results of Image Restoration}
\label{sec:visual}

\paragraph{Image Deblocking.}
We report the additional quantitative results for image deblocking under
different quality factors (20, 30, and 40) on standard benchmarks in ~\cref{tab:supp_deblock}, which compares our method against several state-of-the-art LUT-based and DNN-based methods, including SR-LUT~\cite{srlut}, MuLUT~\cite{mulut}, SPF-LUT~\cite{spflut}, as well as DNN-based approaches such as ARCNN~\cite{arcnn}, and SwinIR~\cite{swinir}.

\paragraph{Image Denoising.}
We report the additional quantitative results for
grayscale image denoising at a noise level of 25 and 50 on
standard benchmarks in ~\cref{tab:supp_denoise}, which compares our method against several state-of-the-art LUT-based and DNN-based methods, including SR-LUT~\cite{srlut}, MuLUT~\cite{mulut}, SPF-LUT~\cite{spflut}, as well as DNN-based approaches such as DnCNN~\cite{dncnn}, FFDNet~\cite{zhang2018ffdnet} and SwinIR~\cite{swinir}.

\begin{table}[h]
\centering
\renewcommand{\arraystretch}{1.1}
\setlength{\tabcolsep}{1mm}
\caption{PSNR results of different methods on Set12 and BSD68 (noise levels 25, 50).}
\label{tab:supp_denoise}
\begin{tabular}{llcccc}
\hline
\textbf{PSNR} & \textbf{Method} & \multicolumn{2}{c}{\textbf{Set12}} & \multicolumn{2}{c}{\textbf{BSD68}} \\
 & & 25 & 50 & 25 & 50 \\
\hline
\multirow{4}{*}{LUT}
  & SR-LUT         & 27.19 & 22.62 & 26.85 & 22.39 \\
  & MuLUT            & 28.94 & 25.46 & 28.18 & 24.97 \\
  & SPF-LUT+DFC          & 29.43 & 26.07 & 28.50 & 25.50 \\
  & \textbf{ShiftLUT-F(ours)}       & \textcolor{red}{29.98} & \textcolor{red}{26.76} & \textcolor{red}{28.90} & \textcolor{red}{25.98} \\
\hline
\multirow{3}{*}{DNN}
  & DnCNN     & 30.44 & 27.18 & 29.23 & 26.23 \\
  & FFDNet   & 30.54 & 27.32 & 29.19 & 26.29 \\
  & SwinIR   & 31.01 & 27.97 & 29.50 & 26.58 \\
\hline
\end{tabular}
\end{table}

\section{Pseudo Code of EAS}
\label{sec:code_eas}

Algorithm \ref{alg:eas_pipeline} provides the detailed pseudo-code for the proposed Error-bounded Adaptive Sampling (EAS) pipeline, which encompasses both the offline optimization and online inference stages.

\begin{algorithm}[t]
\small
\caption{EAS Pipeline: Optimization \& Inference}
\label{alg:eas_pipeline}
\begin{algorithmic}[1]
\State \textbf{Stage 1: Offline Optimization}
\State \textbf{Input:} Pre-trained LUTs $\{\mathcal{L}_1, \dots, \mathcal{L}_L\}$, tolerance $\varepsilon$
\For{each layer $l = 1 \dots L$}
    \State Search for max stride $s_l^* \in \mathcal{S}$ s.t. $\mathrm{Error}(s_l^*) \le \varepsilon$
    \State Store compressed $\mathcal{L}_{l}^{sparse}$
\EndFor
\vspace{-0.5em}
\Statex \hrulefill

\State \textbf{Stage 2: Online Inference} (Serial Cascade)
\State \textbf{Input:} Image $x$, Compressed $\{\mathcal{L}_1^{sparse}, \dots\}$, Shared Buffer $\mathcal{B}$
\For{each layer $l = 1 \dots L$} \Comment{Process layers sequentially}
    \State \textit{// Step A: Refresh Buffer with current layer's LUT}
    \For{$i \in 0 \dots 2^{\text{MSB}}-1$}
        \State $\mathcal{B}[i] \leftarrow \mathrm{Query}(\mathcal{L}_{l}^{sparse}, i, s_l^*)$ \Comment{Pre-compute}
    \EndFor
    
    \State \textit{// Step B: Pixel-wise Inference (Zero-overhead lookup)}
    \For{each pixel $p$ in $x$}
        \State $x_p \leftarrow x_p + \mathcal{B}[x_p]$ \Comment{Residual lookup using buffer}
    \EndFor
\EndFor
\State \Return Processed feature map $x$
\end{algorithmic}
\end{algorithm}
